\documentclass[remotesensing,accept,article,moreauthors,pdftex,10pt,a4paper]{mdpi} 

\firstpage{1} 
\articlenumber{x}
\doinum{10.3390/------}
\pubvolume{xx}
\pubyear{2016}
\copyrightyear{2016}
\externaleditor{Academic Editor: Qi Wang}
\history{Received: date; Accepted: date; Published: date}

\pdfoutput=1

\usepackage{amsmath,amsfonts,amssymb}
\usepackage{hyperref}
\usepackage{multirow}
\usepackage{bbm}
\usepackage[tight]{subfigure}
\usepackage{siunitx}
\usepackage{booktabs}

\usepackage{tikz}
\usetikzlibrary{positioning}

\usepackage{multicol}

\Title{UNASSISTED QUANTITATIVE EVALUATION OF DESPECKLING FILTERS}

\Author{Luis Gomez $^{1,*}$, Raydonal Ospina $^{2}$ and Alejandro~C.~Frery $^{3}$}
\AuthorNames{Luis Gomez, Raydonal Ospina and Alejandro C.\ Frery}

\address{%
$^{1}$ \quad Department of Electronic Engineering and Automation, University of Las Palmas de G.C., Spain; luis.gomez@ulpgc.es\\
$^{2}$  \quad Departamento de Estat\'istica.  Universidade Federal de Pernambuco, Brazil; rayospina@gmail.com \\
$^{3}$  \quad LaCCAN -- Laborat\'orio de Computa\c c\~ao Cient\'ifica e An\'alise Num\'erica. Universidade Federal de Alagoas, Brazil; acfrery@laccan.ufal.br
} 

\corres{Correspondence: luis.gomez@ulpgc.es}

\firstnote{The authors contributed equally to this work.} 

\abstract{SAR (Synthetic Aperture Radar) imaging plays a central role in Remote Sensing due to, among other important features, its ability to provide high-resolution, day-and-night and almost weather-independent images.
SAR images are affected from a granular contamination, speckle, that can be described by a multiplicative model. 
Many despeckling techniques have been proposed in the literature, as well as measures of the quality of the results they provide. 
Assuming the multiplicative model, the observed image $Z$ is the product of two independent fields: the backscatter $X$ and the speckle $Y$. 
The result of any speckle filter is $\widehat X$, an estimator of the backscatter $X$, based solely on the observed data $Z$. 
An ideal estimator would be the one for which the ratio of the observed image to the filtered one $I=Z/\widehat X$ is only speckle: a collection of independent identically distributed samples from Gamma variates. 
We, then, assess the quality of a filter by the closeness of $I$ to the hypothesis that it is adherent to the statistical properties of pure speckle. 
We analyze filters through the ratio image they produce with regards to first- and second-order statistics: the former check marginal properties, while the latter verifies lack of structure. 
A new quantitative image-quality index is then defined, and applied to state-of-the-art despeckling filters. 
This new measure provides consistent results with commonly used quality measures (equivalent number of looks, PSNR, MSSIM, $\beta$ edge correlation, and preservation of the mean), and ranks the filters results also in agreement with their visual analysis.
We conclude our study showing that the proposed measure can be successfully used to optimize the (often many) parameters that define a speckle filter.}


\begin{document}



\section{Introduction}
\label{Sec:Introduction}

Speckle reduction has occupied both the scientific literature and the production software industry since the deployment of SAR platforms. 
Good speckle filters are expected to improve the perceived image quality while preserving the scene reflectivity.
The former requires, at the same time, preservation of details in heterogeneous areas and constancy in homogeneous targets.

Early works assessed the performance of despeckling techniques by visual inspection of the filtered images; cf.\ Refs.~\cite{ImageEnhancementNoiseFilteringLocalStatistics,Lee86}.
Since then, speckle filtering has reached such a level of sophistication~\cite{art:Agenti_2013} that forthcoming improvements are likely to be incremental, and assessing them quantitatively is, at the same time, desirable and hard.
Also, as filters are often defined with many parameters, eg.\ window size, thresholds, etc., finding an optimal setting is also an issue.

The Equivalent Number of Looks ({\text{ENL}}) is among the simplest and most spread measures of quality of despeckling filters.
It can be estimated, in textureless areas and intensity format, as the ratio of the squared sample mean to the sample variance, i.e., the reciprocal of the squared coefficient of variation (see~\cite{oliverquegan98} for other methods for the estimation of {\text{ENL}}). 
Being proportional to the signal-to-noise ratio, the higher {\text{ENL}} is, the better the quality of the image is in terms of speckle reduction. 
However, it is well known that large {\text{ENL}} values are easily obtained just by overfiltering an image, which severely degrades details and gives the filtered image an undesirable blurred appearance.
In particular, $\text{ENL}=\infty$ is obtained in completely flat areas where the sample variance is null.
Testing a filter merely by its performance over textureless areas, where a simple generic filter as the Boxcar, would perform well, is bound to produce misleading results.  

Other measures of quality commonly used for speckle filter assessment enhance certain characteristics, but suffer from shortcomings. 
The proposal and assessment of a new filter is frequently supported by a plethora of measures.
As such, it is hard to used them to optimize the parameters that often specify a filter.

An alternative approach for assessing the performance of despeckling methods is the analysis of ratio images, as proposed in~\cite{oliverquegan98}. 
This is becoming a standard procedure in the SAR community~\cite{art:Achim_2006, art:Parrilli_2012, art:Martino_2014, art:Gomez_JSTARS_2015}.
It consists of checking by visual inspection whether patterns appear in the ratio image $I = Z/\widehat{X}$, where $Z$ is the original image and $\widehat{X}$ is its filtered version. 
Under the multiplicative model, the ratio image from the ideal filter should be pure speckle with no visible patterns.
The presence of geometric structures, changes of statistical properties, or any detail correlated to the original image $Z$ in $I$ indicates poor filter performance, i.e., not only speckle but also other possible relevant information has been removed from the original image.
The visual interpretation of ratio images, being subjective, is qualitative and irreproducible.

\begin{figure}[hbt]
\centering
\includegraphics[width=.4\linewidth]{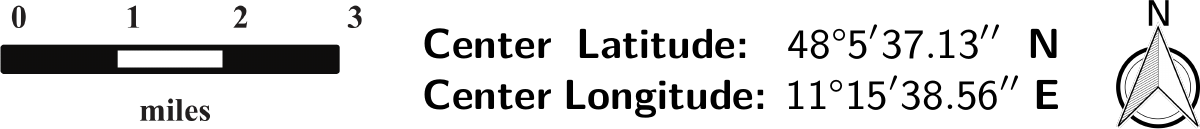}\\
\includegraphics[width=.4\linewidth]{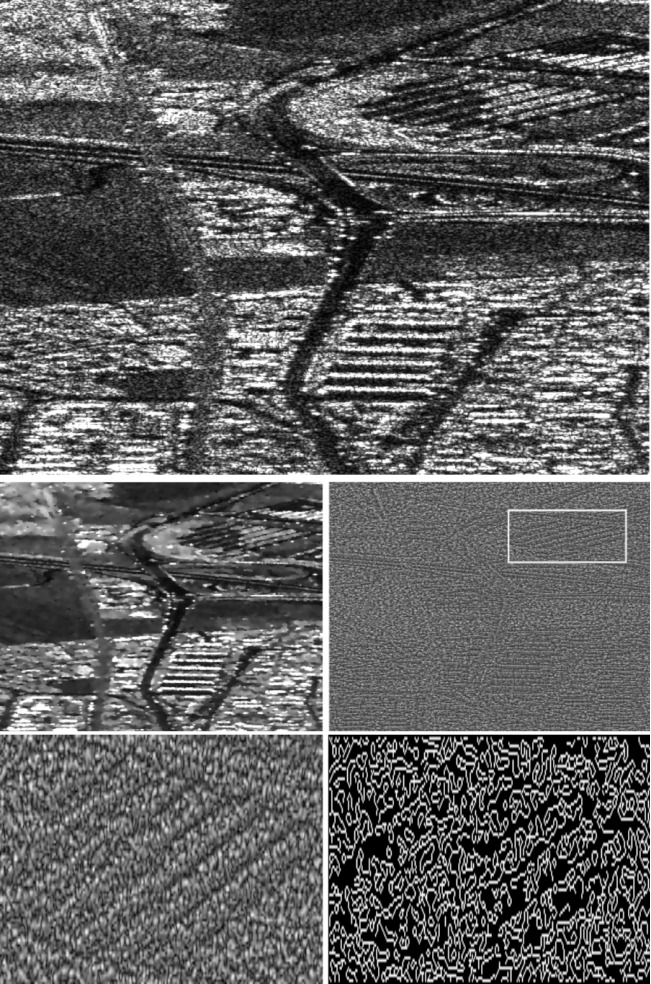}
	\caption{Top: original SAR image. Middle: SRAD  ($ T = 50 $) filtered image and ratio image. Bottom: zoom of a selected area within the ratio image and extracted edges by Canny's edge detector.}
\label{fig:Sub_SAR_photo_1_all}
\end{figure}

Fig.~\ref{fig:Sub_SAR_photo_1_all} illustrates this idea. 
This image is part of a single-look HH SAR data set obtained over Oberpfaffenhofen, Germany, with textureless areas, bright scatterers, and urban areas with geometric content as buildings and roads. 
Fig~\ref{fig:Sub_SAR_photo_1_all} (top) is the original speckled image,
and below left is its filtered version obtained with the SRAD (speckle anisotropic diffusion) filter~\cite{art:Yu_2002}. 
The filtered image is acceptable in terms of edge and details preservation: textureless areas look smooth, as expected after a successful despeckling.
The middle row right is the resulting ratio image, with a ROI (region of interest) in the urban area. 
The third row of Fig.~\ref{fig:Sub_SAR_photo_1_all} (left) presents a zoom of the highlighted area. 
It shows remaining structures in the ratio image, an evidence that the SRAD filter is not ideal for this case.

The quantitative assessment of such residual geometrical content is a challenging task because, besides being subtle, it has similar properties to the rest of the ratio image: brightness, marginal distribution etc. 
That is, areas with and without geometrical structure (even narrow edges) are extremely noisy and, therefore, simple algorithms as, for instance, those based on edge detection, fail at detecting them; cf. the result of applying the Canny edge detector in the third row of Fig.~\ref{fig:Sub_SAR_photo_1_all}, right. 
Also, the better the filter is, the harder will be identifying and quantifying remaining structures in the ratio image.

This work proposes a new measure of quality that does not require any ground reference.
Using only the original image, an estimate of its number of looks, and the filtered image, we measure the deviation from the ideal filter as a combination of deviations from the ideal marginal properties with a measure of remaining structure in the ratio image.
We test this unassisted measure of quality in both simulated data and on images obtained by an actual SAR sensor, and we show it is able to rank with a single value the results produced by four state-of-the-art filters in a way that captures other measures of quality.
We also show it can be used to fine-tune filter parameters.

The remainder of this article is organized as follows.
Section~\ref{Sec:SARImageRatio} recalls the basic assumptions underlying this proposal: the multiplicative model.
With this in view, we discuss the properties to be measured in a ratio image.
Section~\ref{Sec:Methodology} presents our proposal of an unassisted quantitative measure for assessing the quality of despeckling filters.
In Sec.~\ref{Sec:Experimental Setup and Results} we present the results observed on both simulated and actual SAR images, and show an example of filter parameter tuning.
Section~\ref{Sec:Conclusion} concludes the article.

\section{SAR image formation and ratio images}
\label{Sec:SARImageRatio}

Although we recognize the nature of SAR data depends of many system parameters, our work starts by assuming the multiplicative model for the observations.
Observations can be, thus, described by the product of two independent variables, $X$ and $Y$ that model, respectively, the (desired but unobserved) backscatter and the speckle noise.
So, $Z=XY$ models the observed data, and one aims at obtaining $\widehat{X}$, a good estimator of $X$.

Without loss of generality, we will assume the available data is in intensity format, i.e., power.
Amplitude data should be squared before applying our method.

The usual assumption is that $Y$ is a collection of independent identically distributed Gamma random variables with unitary mean, and shape parameter equal to the number of looks.
The backscatter is constant in textureless areas, and otherwise can be described by another random variable.

Our main aim is assessing the quality of despeckling filters by measuring how the ratio images they produce deviate from the idealized result.

The perfect filtered image is $\widetilde{X}=X$ and, thus, produces a ratio image $Z/\widetilde{X}=Y$ which consists of pure speckle.
Based on this observation, our measure of quality captures departures from the following hypothesis:
``the perfect speckle filter leads to a ratio image formed by a collection of independent identically distributed Gamma random variables with unitary mean and shape parameter equal to the (equivalent) number of looks the original image has''.

In the following, we illustrate our idea with images and one-dimensional slices.
We elaborate three situations to make our point on the usefulness of ratio images for detecting the performance of a speckle filter.

Firstly, we will see the effect of oversmoothing textured areas.

\begin{figure}[hbt]
\centering
\subfigure[Constant step and speckled return\label{fig:ConstantStep}]{\includegraphics[width=.48\linewidth]{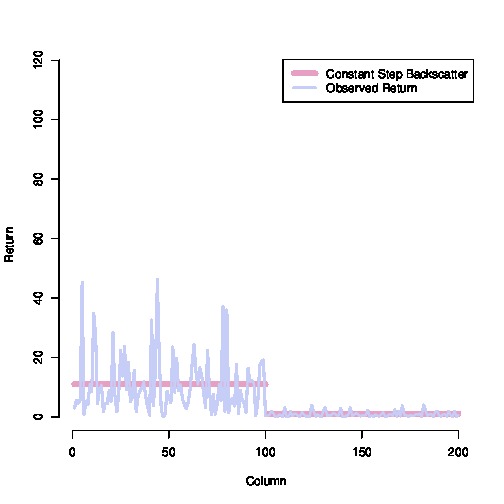}}
\subfigure[Textured step and speckled return\label{fig:TexturedStep}]{\includegraphics[width=.48\linewidth]{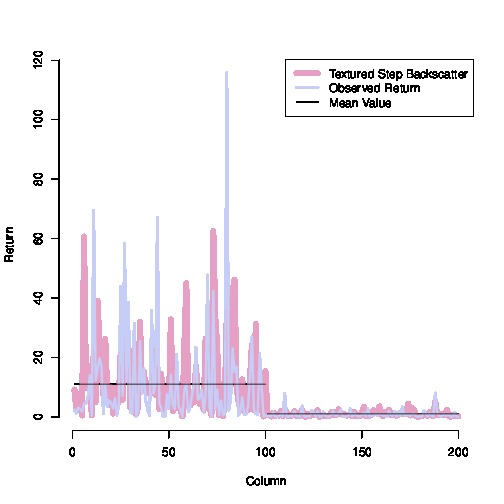}}
\caption{A step: constant and textured versions, and their return.}
\label{fig:Step}
\end{figure}

Figure~\ref{fig:ConstantStep} shows a step function in pink (the backscatter), and the observed return from this backscatter in single look fully developed speckle, i.e., exponential deviates with mean equal to $11$ (left half) and $1$ (right half).

Figure~\ref{fig:TexturedStep} shows a similar situation, but when the backscatter is no longer constant.
In this case, the backscatter is textured with mean $11$ and $1$, as in the previous example, but varying according to exponential deviates.
The textured step backscatter is shown in pink.
When speckle enters the scene, modeled here again as unitary mean exponential random variables, the observed data obeys a $\mathcal K$ distribution; shown in lavender.

What should the ideal filter return?
It is our understanding that $\widetilde X$ should be the underlying backscatter, i.e.,
either the step function in the case where there is no texture, 
or the textured observations without speckle (both depicted in pink in Fig.~\ref{fig:Step}).

\begin{figure}[hbt]
\centering
{\includegraphics[width=.5\linewidth]{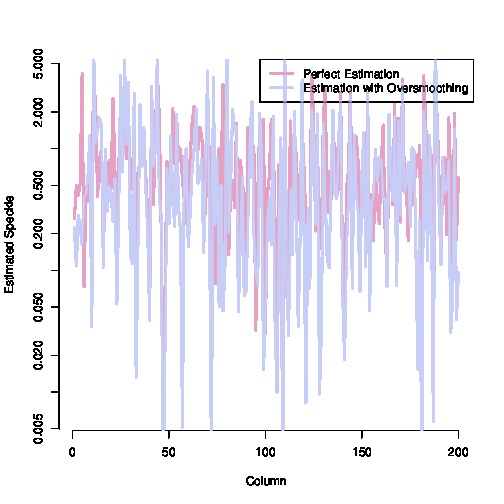}}
\caption{Estimated speckle by the ideal filter and by overmoothing.}
\label{fig:EstimatedSpeckle}
\end{figure}

A filter that returns the step function in the textured case (thin black line in Fig.~\ref{fig:TexturedStep}) is oversmoothing.
Fig.~\ref{fig:EstimatedSpeckle} shows, in semilogarithmic scale, the estimated speckle as produced by the ideal filter (pink) and by oversmoothing (lavender); these are the resulting ratio images from the ideal and a poor filter, respectively.
This last estimate is the result of dividing the observed return from Fig.~\ref{fig:TexturedStep} by the step function.

The effect of oversmoothing is noticeable: the speckle produced by the ideal filter has less variability than the one resulting from returning the step function as estimator.
While the sample variance of pure speckle is $s^2=0.80$, that of the speckle with remaining structure is $s^2=2.12$.
Although numerically detectable by first-order statistics, this effect is seldom visible.

Secondly, we will see how neglecting structures impacts on ratio images.

\begin{figure}[hbt]
\centering
\subfigure[Slowly-varying mean value and its return\label{fig:SlowlyVarying}]{\includegraphics[width=.48\linewidth]{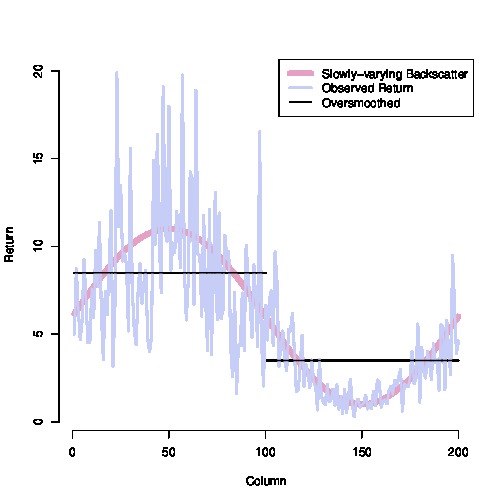}}
\subfigure[Estimated speckle\label{fig:EstimSpeckleSlowlyVaying}]{\includegraphics[width=.48\linewidth]{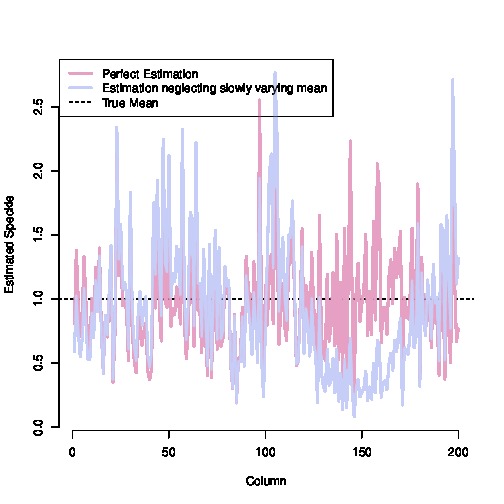}}
\caption{Slowly varying backscatter, fully developed speckle, and estimated speckle.}
\label{fig:VaryingBackscatter}
\end{figure}

Fig.~\ref{fig:VaryingBackscatter} shows the situation of fully developed speckle, in this case with three looks.
It affects an structure seen as slowly-varying backscatter, the sine curve depicted in pink.
The observed return, obtained as the point-by-point product of the speckle with the backscatter is shown in lavender; Fig.~\ref{fig:SlowlyVarying}.

On the one hand if, as we postulate, the ideal filter retrieves the true backscatter, the ratio image or estimated speckle will coincide with the true speckle (in pink in Fig~\ref{fig:EstimSpeckleSlowlyVaying}).
On the other hand, if the filter oversmooths the backscatter and returns a step function (in black in Fig.~\ref{fig:SlowlyVarying}), the resulting estimated speckle will retain part of the missing estructure; cf.\ Fig.~\ref{fig:EstimSpeckleSlowlyVaying} in lavender.

\begin{figure}[hbt]
\centering
{\includegraphics[width=.3\linewidth]{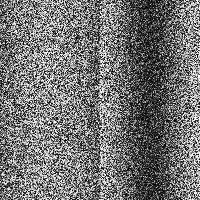}}
\caption{Ratio image resulting from neglecting a slowly varying structure under fully developed speckle.}
\label{fig:ResidualSine}
\end{figure}

Figures~\ref{fig:EstimatedSpeckle} and~\ref{fig:EstimSpeckleSlowlyVaying} also show that detecting departures from the ideal situation is a hard task.
Figure~\ref{fig:ResidualSine} shows how the ratio image obtained from neglecting the slowly varying structure looks like.
We postulate and show evidence that this remaining structure can be effectively detected and quantified with second-order statistics.

Finally, we will see how a poor filter will render a ratio image with detectable structure when dealing with edges.

\begin{figure}[hbt]
\centering
\subfigure[Strips and speckle\label{fig:StripsSpeckle}]{\includegraphics[width=.48\linewidth]{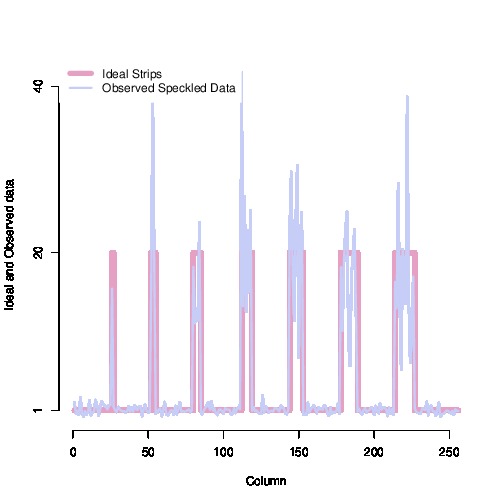}}
\subfigure[Filtered strips with oversmoothing\label{fig:FilteredStrips}]{\includegraphics[width=.48\linewidth]{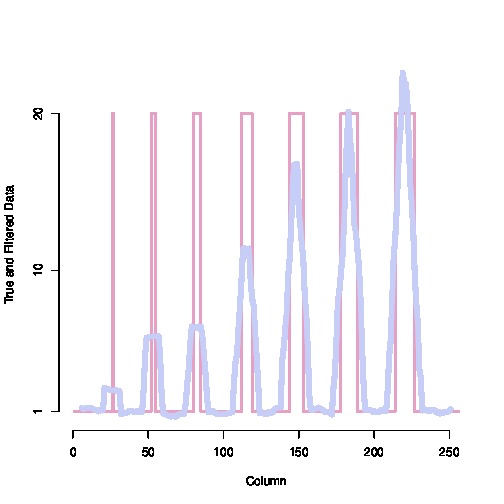}}
\caption{The effect of oversmoothing on an image of strips of varying width.}
\label{fig:Strips}
\end{figure}

Fig.~\ref{fig:StripsSpeckle} shows a line of the strips image typical of articles that analyze the performance of speckle filters with simulated data; cf.~\cite{art:Lee_1994,DespecklingPixelRelativity}.
The strips take two values: $1$ and $20$ (pink), the speckle is a collection of i.i.d. Gamma variates with three looks and unitary mean, and the observed data (in lavender) is the product of the strips and speckle.

Fig.~\ref{fig:FilteredStrips} shows, again, the strips and the estimated backscatter as returned by a simple filter: the local mean using eleven observations.
The oversmoothing is noticeable.
It not only degrades the sharpness of the edges, but also reduces observed value.
This last effect is more noticeable over narrow strips (to the left of the figure).

\begin{figure}[hbt]
\centering
{\includegraphics[width=.5\linewidth]{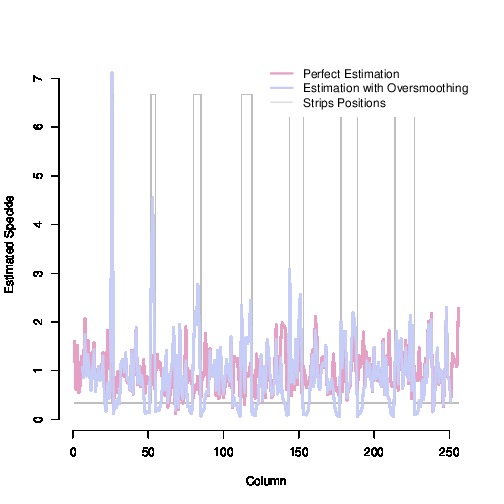}}
\caption{Estimated speckle: ideal and oversmoothing filters.}
\label{fig:EstimatedStripsSpeckle}
\end{figure}

The estimated speckle, as expected, will be affected by the poor result returned by the local mean filter, as shown in Fig.~\ref{fig:EstimatedStripsSpeckle}.
The true speckle is shown in pink, while the one estimated using the oversmoothed backscatter tends to have peaks where the smaller strips are (cf. the lavender signal).
This will affect the ratio images rendering data whose behavior departs from the ideal situation, which is a collection of i.i.d. deviates from the a Gamma distribution with unitary mean and shape parameter equal to the equivalent number of looks of the original image.

\begin{figure}[hbt]
\centering
\subfigure[Speckled strips\label{fig:StripsImageSpeckled}]{\includegraphics[width=.32\linewidth]{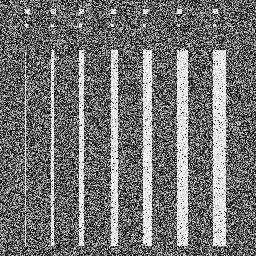}}
\subfigure[Filtered strips\label{fig:FilteredImageStrips}]{\includegraphics[width=.32\linewidth]{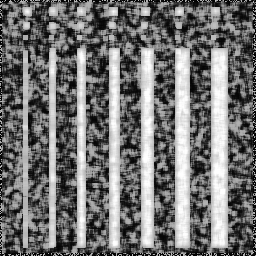}}
\subfigure[Ratio image\label{fig:RatioImageStrips}]{\includegraphics[width=.32\linewidth]{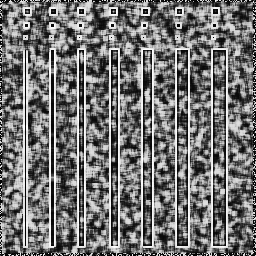}}
\caption{Speckled strips, result of applying a $5\times 5$ BoxCar filter, ratio image.}
\label{fig:StripsImages}
\end{figure}

Figure~\ref{fig:StripsImages} shows these effects in the strips image.
Again, we postulate that identifying and quantifying the departure from the ideal filter, i.e., the remaining structure visible in Fig.~\ref{fig:RatioImageStrips}, is feasible with both first- and second-order statistics.

\section{Unassisted measure of quality based on first- and second-order descriptors}
\label{Sec:Methodology}

We propose an evaluation based on two components. 
A statistical measure of the quality of the remaining speckle is the first-order component of the quality measure.
This component is comprised of two terms: one for mean preservation, and another for preservation of the equivalent number of looks
The second-order component measures the remaining geometrical content within the ratio image. 
The three elements that comprise our measure of quality are relative, in order to make them comparable.

As pointed out before, the usual approach to evaluate ratio images consists of, after the visual inspection,  to  estimate the {\text{ENL}} within an homogeneous area.
Then, the best filter is the one for which the ratio image has the mean value closest to unity and the equivalent number of looks closest to the {\text{ENL}} of the original (noisy) image (see for instance~\cite{art:Achim_2006}).

To avoid user intervention, which is one of the requirements of our proposal, we automatically select suitable textureless areas.
First, we estimate the local mean and standard deviation on sliding windows of side $w$ over the original image.
With these values, we compute the local ENL ($\widehat{\text{ENL}}_{\text{noisy}}$) as the reciprocal of the squared coefficient of variation.
Then, we also compute the local mean and standard deviation on the ratio image with the same window, obtaining $\widehat\mu_{\text{ratio}}$ and $\widehat{\text{ENL}}_{\text{ratio}}$.

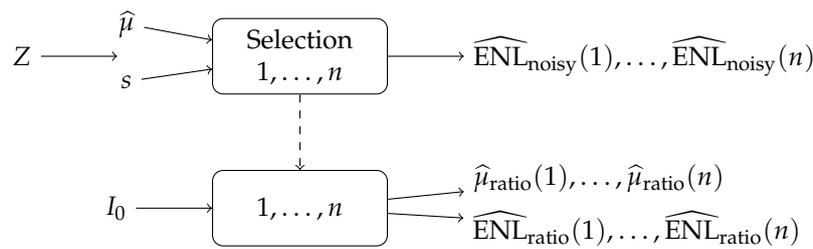
\begin{figure}[hbt]
\centering
\begin{tikzpicture}
[caixa/.style={rectangle,draw,minimum height=10mm,minimum width=23mm,rounded corners}]

\node(caixa1)[caixa, inner sep=1mm, text width=4em, text centered]{{Selection} $1,\dots, n$};
\node(phan)[left =of caixa1]{ };
\node(muh)[above =of phan, yshift=-1cm] {$\widehat\mu$ };
\node(s)[below =of phan, yshift=1cm] {$s$};
\node(Z)[left =of phan]{$Z$};
\node(caixa2)[caixa][below=of caixa1, yshift=0cm]{$1,\dots, n$};
\node(I0)[left=of caixa2] {$I_0$};

\node(ENL)[right=of caixa1] {$\widehat{\text{ENL}}_{\text{noisy}}(1),\dots,\widehat{\text{ENL}}_{\text{noisy}}(n)$};

\node(mu2)[right=of caixa2,  yshift=0.4cm] {$\widehat{\mu}_{\text{ratio}}(1),\dots,\widehat{\mu}_{\text{ratio}}(n)$};

\node(ENL2)[right=of caixa2, below=of mu2,  yshift=1.0cm, xshift=0.5cm] {$\widehat{\text{ENL}}_{\text{ratio}}(1),\dots,\widehat{\text{ENL}}_{\text{ratio}}(n)$};

\draw[->] (Z) -- (phan);
\draw[->] (muh) -- (caixa1);
\draw[->] (s) -- (caixa1);
\draw[->] (I0) -- (caixa2);
\draw[->] (caixa1) -- (ENL);
\draw[->] (caixa2) -- (mu2);
\draw[->] (caixa2) -- (ENL2);
\draw[->,dashed] (caixa1) -- (caixa2);
\end{tikzpicture}
\caption{Selection of mean and ENL values for the first-order measure.}
\label{fig:FirstOrder}
\end{figure}

We select as textureless areas those where both $\widehat{\text{ENL}}_{\text{ratio}}$ is close enough to $\widehat{\text{ENL}}_{\text{noisy}}$ and $\widehat\mu_{\text{ratio}}$ is close enough to $1$.
We stipulate a tolerance for the absolute relative error, and with this we select $n$ areas.
This procedure is illustrated in Fig.~\ref{fig:FirstOrder}.

Then, for the $n$ selected homogeneous areas, we calculate the first-order residual as  
\begin{equation}
r_{\widehat{{\text{ENL}}}, \widehat{\mu}} = \frac{1}{2}\sum_{i = 1}^{n}{\left(r_{\widehat{{\text{ENL}}}}(i) + r_{\widehat{\mu}}(i)\right)},
\label{eq:total_residue_estimator_mean}
\end{equation}
where, for each homogeneous area $i$,
$$
r_{\widehat{{\text{ENL}}}}(i) = \frac{|\widehat{{\text{ENL}}}_{\text{noisy}}(i) - \widehat{{\text{ENL}}}_{\text{ratio}}(i)|}
{\widehat{{\text{ENL}}}_{\text{noisy}}(i)}
$$
is the absolute value of the relative residual due to deviations from the ideal ENL, and
$$
r_{\widehat{\mu}}(i) = | 1 - \widehat{\mu}_{\text{ratio}}(i)|
$$ 
is the absolute value of the relative residual due to deviations from the ideal mean (which is $1$).
An ideal despeckling operation would yield $r_{\widehat{{\text{ENL}}}, \widehat{\mu}}  = 0$.

We measure the remaining geometrical content with the inverse difference moment (also called \textit{homogeneity}) from Haralik's co-ocurrence matrices~\cite{TexturalFeaturesImageClassification,BenchmarkEUSAR2016}.
Low values are associated with low textural variations and vice versa.
Let $P(i,j)$ be a co-occurence matrix at an arbitrary position, and $p(i,j)= P(i,j)/K$ its normalized version, with $K$ a constant.
The homogeneity, our second-order measure, is
\begin{equation}
h = \sum_i{\sum_j{\frac{1}{1 + (i - j)^2}\cdot p(i,j)}}.
\label{eq:homogeneity}
\end{equation}
This is computed for every coordinate, yielding measures of the remaining structure, but we need a reference to compare it with.

The null hypothesis implies that the probability distribution of the ratio image $I$ is invariant under random permutations, i.e., if $I_1,I_2,\dots,I_M$ are independent identically distributed random variables, also are $g(I_1,I_2,\dots,I_M)$, any random permutation.
Applying this idea, we measure the geometric content in a ratio image evaluating $h$ on the ratio image and then on a shuffled versions of it.
If there is no structure in $I$, $h$ will not change after shuffling, but if $I$ has structure, then shuffling will tend to destroy it.

Let $h_{\text o}$ and $h_g$ be the mean of all values of homogeneity obtained from the original ratio image $I_{\text o}$ and from the result of randomly permuting all its values $I_{g}$, respectively. 
We use $\delta h =100|h_{\text o}-\overline{h_g}| / h_{\text o}$, the absolute value of the relative variation of $h_{\text o}$ in percentage as a measure of the departure from the null hypothesis: the larger this variation is, the greater the amount of structure relies on the ratio image.
Here $\overline{h_g}$ is the average over $p\geq 1$ samples of $I_{g}$.

Since the spatial structure is subtle in ratio images produced by state-of-the-art filters, $\delta h$ requires being scaled to be comparable with $r_{\widehat{{\text{ENL}}}}$.
After careful experimentation with both simulated data and images from operational sensors, we found that $100$ produces sensible and consistent results.
This value was then fixed as part of our proposal, requiring no further tuning.
Note that $\delta h$ provides an objective measure for ranking despeckled results regarding solely the remaining geometrical content within the related ratio images.

The proposed estimator combines the measures of the remaining structure and of deviations from the statistical properties of the ratio image:
\begin{equation}
 \mathcal{M} = r_{\widehat{{\text{ENL}}}, \widehat{\mu}} + \delta h,
 \label{eq:M}
\end{equation}
The perfect despeckling filter will produce $\mathcal{M} = 0$, and the larger $\cal M$ is, the further the filter is from the ideal.

In the following, we will show that the proposed measure of quality is expressive and able to translate into a single value a number of measures of quality, both objective and subjective.

\section{Experimental Setup}
\label{Sec:Experimental Setup and Results}

In this section we present the results of using the new metric for evaluating the quality of widely-used despeckling filters.  
We employ both simulated data and images from operational SAR systems, and we conclude with an application of our metric for filter optimization.

We used the following filters: 
E-Lee (Enhanced Lee~\cite{art:Lee_2009}),
SRAD (Speckle Reducing Anisotropic Diffusion~\cite{art:Yu_2002}), 
PPB (Probabilistic Patch Based~\cite{art:Deledalle_2009}), 
and FANS (Fast Adaptive Nonlocal SAR~\cite{art:Cozzolino_2014}). 
All of them provide good results and may be considered state-of-the art despeckling filters. 
E-Lee filter is an improved version of the classical adaptive Lee filter~\cite{Lee86}. 
SRAD belongs to the category of PDE-based (Partial Differential Equations) filters, while the other two belong to the category of nonlocal means filters.
In particular, FANS employs a set of wavelet transforms in its collaborative filtering stage.

The filters were tuned to the recommended designs as provided by their authors, with slight modifications (mask size and related threshold values) for PPB and FANS that yielded improved mean and ENL preservation. 
This was done for a fair comparison with SRAD and E-Lee filters which perform particularly well on preserving those features.

The E-Lee filter uses a $9\times9$ search window, and all the other parameters are as in~\cite{art:Lee_2009}.
The diffusion time for SRAD is $T=300$, and the other parameters are as recommended in~\cite{art:Yu_2002}. 
The PPB filter uses $7\times7$ patches and $21\times21$ search windows, and $25$ iterations.  
The FANS filter uses $8\times8$ blocks, and $39\times39$ pixels search area; the remaining parameters are set as specified in~\cite{art:Cozzolino_2014}.  
The E-Lee and the SRAD filters are our own implementation. 
The source codes of PPB and FANS are available at~\cite{web_PPB} and~\cite{web_FANS}, respectively. 

For all the experiments, 
the co-occurrence matrices were computed after quantizing the observations to eight values,
$p=100$ independent samples were obtained for each image,
and 
the tolerance and window side for eq.~\eqref{eq:total_residue_estimator_mean}
were set to $0.03$ and $w=25$, respectively.
The window side does not have a strong impact on the proposed measure; smaller windows will detect larger textureless patches with less observations, while larger windows will produce the opposite effect.

We will show that usual measures of quality are unable to provide enough evidence for the choice of a filter and, oftentimes, these quantities are conflicting
in both simulated data and images from a SAR sensor.
We will also see that our proposed measure is able to provide a sensible score of filter performance, and to guide in the choice of optimal parameters.

\subsection{Simulation Results}
\label{Simulation_Results}

\begin{figure}[hbt]
	\centering
\subfigure[Blocks and points phantom\label{fig:PhantomBlocksPoints}]{\includegraphics[width=.35\linewidth]{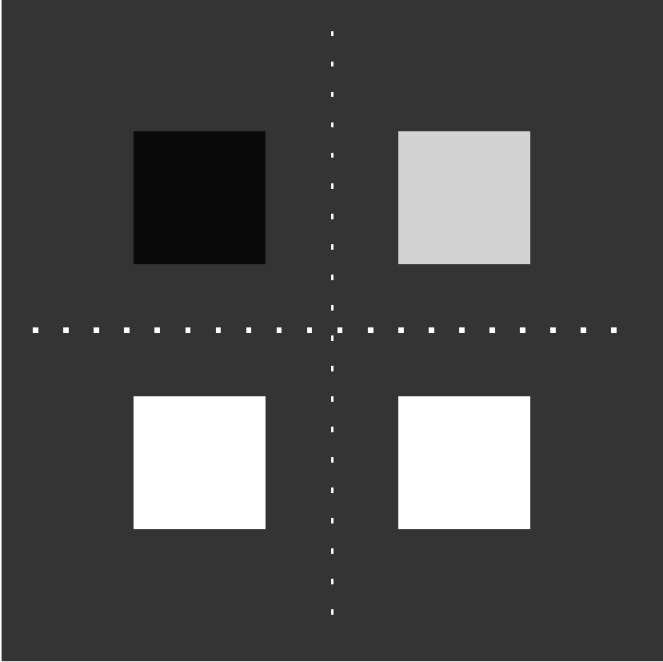}}
\subfigure[Speckled version, single look\label{fig:SpeckleBlocksPoints}]{\includegraphics[width=.35\linewidth]{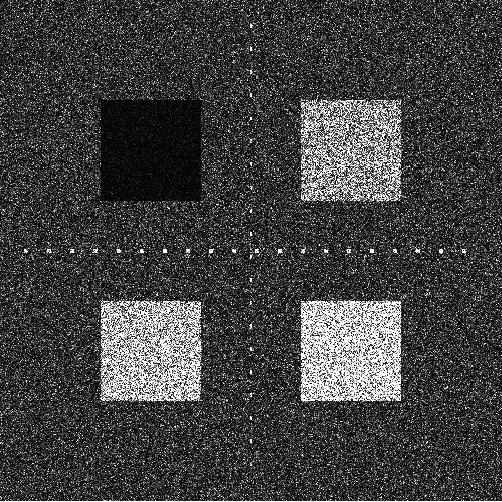}}
			\caption{Blocks and points phantom, and $500 \times500$ pixels simulated single-look intensity image.}
	\label{fig:Phantom_corners}
\end{figure}

Fig.~\ref{fig:PhantomBlocksPoints} shows the phantom with which we simulated images. 
This phantom has both large flat areas, linear edges between them and small pointwise-like details of $2\times4$ and $4\times4$ pixels.
Fig.~\ref{fig:SpeckleBlocksPoints} shows the result of injecting single-look speckle to this phantom.

The data shown in Fig.~\ref{fig:Phantom_corners} allows measuring the ability of speckle filters at reducing noise (it presents large textureless areas), and at preserving small details~\cite{art:Lee_2009,art:Zhong_2011}.
The background intensity is $10$, while that of the four squares is: $2$ (top left), $40$ (top right), $60$ (bottom left), and $80$ (bottom right). 
There are two sets of bright scatterers (intensity $240$): twenty of size $4\times4$ along the horizontal direction, and twenty of size $4\times2$ along the vertical.
The simulated data are obtained by multiplying these values by iid exponential deviates with unitary mean.

\begin{figure}[hbt]
	\centering
		
	{\includegraphics[width=.35\linewidth]{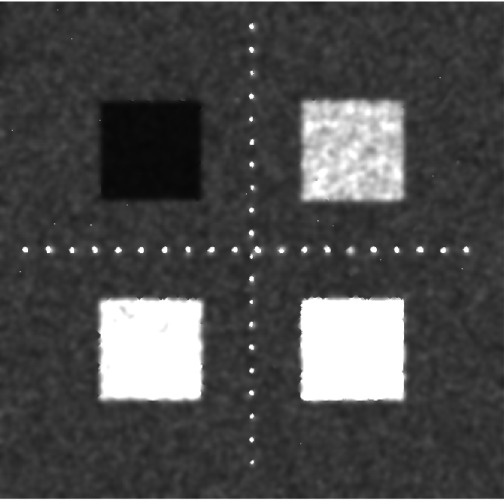}}
	{\includegraphics[width=.35\linewidth]{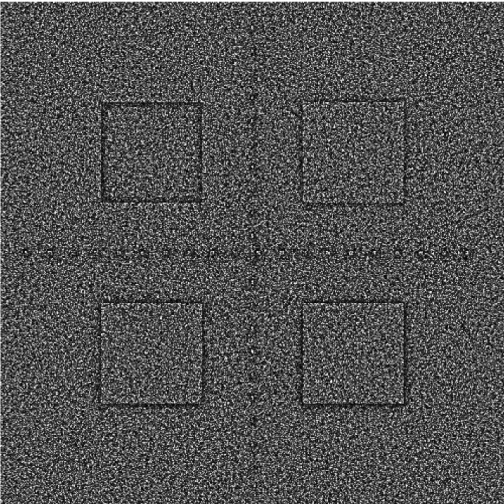}}
	\vskip.5ex
	
	{\includegraphics[width=.35\linewidth]{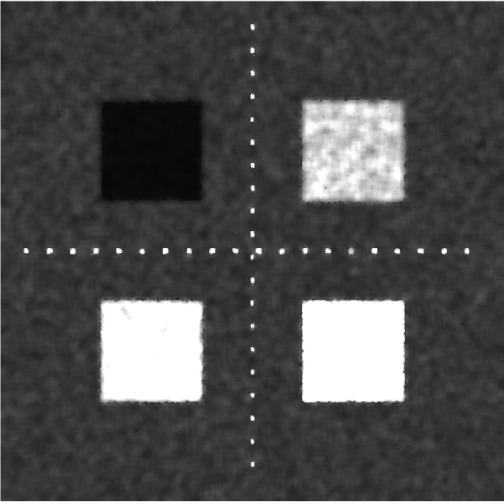}}
	{\includegraphics[width=.35\linewidth]{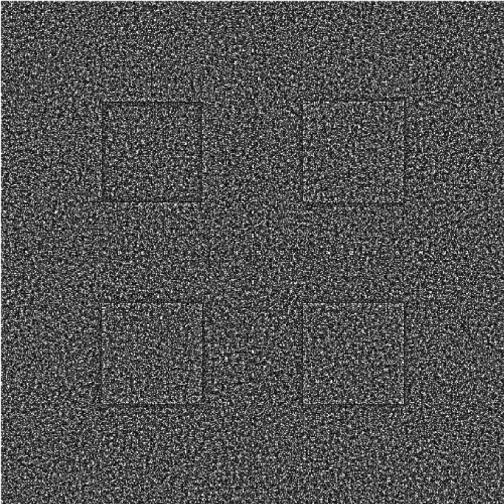}}
	\vskip.5ex
	
	{\includegraphics[width=.35\linewidth]{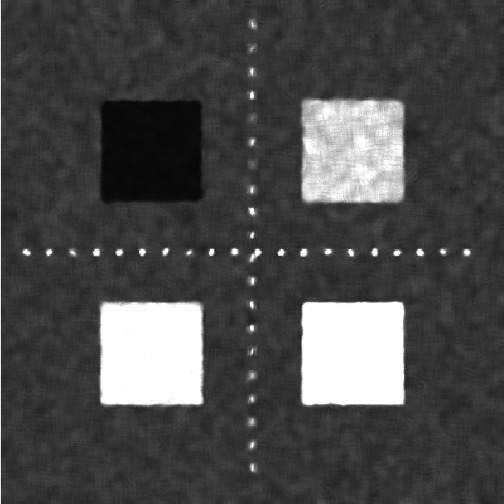}}
	{\includegraphics[width=.35\linewidth]{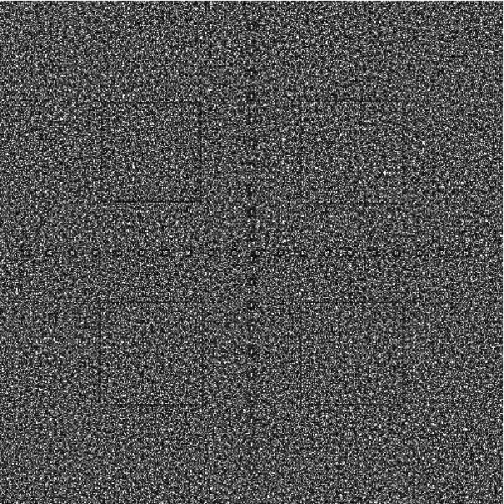}}
	\vskip.5ex
	
	{\includegraphics[width=.35\linewidth]{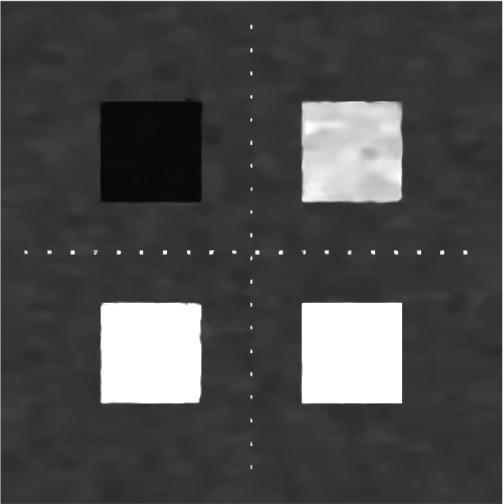}}
	{\includegraphics[width=.35\linewidth]{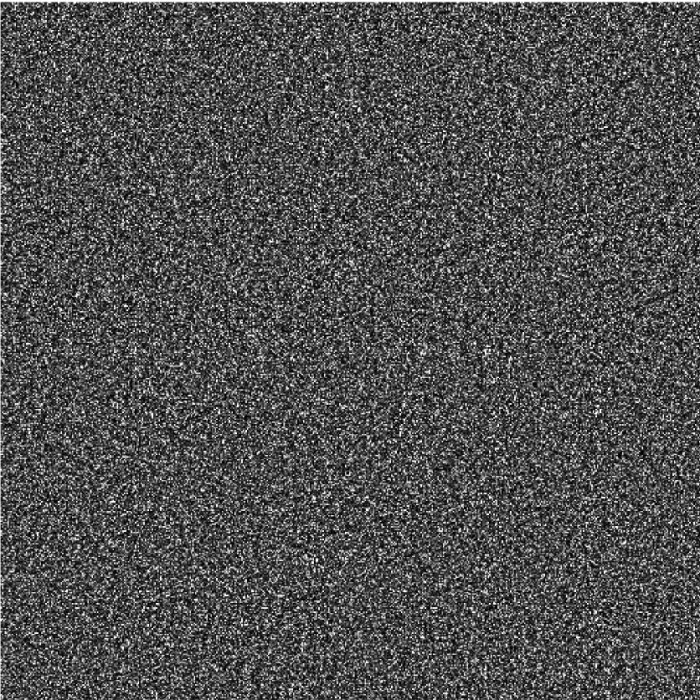}}

	\caption{Results for the simulated single-look intensity data. Top to bottom, (left) results of applying the SRAD, the E-Lee, the PPB and the FANS filters. Top to bottom (right), their ratio images.}
	\label{fig:Phantom_corners_all_SAR_Data_new}
\end{figure}

Fig.~\ref{fig:Phantom_corners_all_SAR_Data_new} shows the results of applying the four filters on the simulated image, and their ratio images (first and second column respectively). 

\begin{figure}[hbt]
	\centering
	{\includegraphics[width=.60\linewidth]{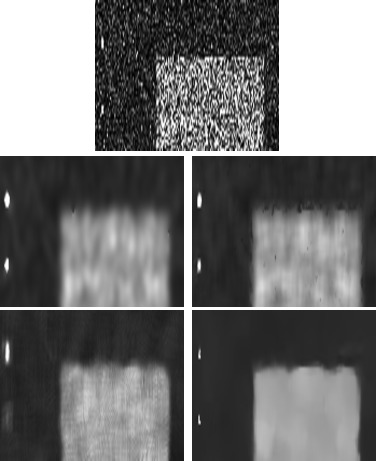}}
			\caption{Zoom of the results for synthetic data: (top) Noisy image, (first row, left) SRAD filter, (first row, right) E-Lee filter, (second row, left) PPB filter and, (second row, right) FANS filter.}
\label{fig:Zoom_Phantom_Corners_All}
\end{figure}

The four filters perform well since they preserve edges and bright scatterers, and also make textureless areas smoother. 
The ratio images reveal that the SRAD, and the E-Lee filters seem to be the least effective in terms of remaining structure as the squares edges are still visible (more for the SRAD filter). 
This remaining geometric content seems minimum for the PPB and FANS filter, although a careful observation reveals structures in all ratio images. 
See details in Fig.~\ref{fig:Zoom_Phantom_Corners_All}.

It is expected that this subjective assessment be confirmed by the quantitative results provided by our proposal. 

An objective assessment can be performed with respect to the ground reference.
To that aim, we computed the Mean Structural Similarity Index MSSIM~\cite{art:Wang_2004}, the Peak Signal-to-Noise Ratio PSNR, and the measure of correlation between edges $\beta$~\cite{art:Sattar_1997}.

MSSIM measures the similarity between the simulated and the despeckled images with local statistics (mean, variance and covariance between the unfiltered and despeckled pixel values)~\cite{art:Chabrier_2008,art:Wang_2004}. 
This measure is bounded in $(-1, 1)$, and a good similarity produces values close to $1$. 
The $\beta$ estimator is useful for assessing edge preservation. 
It evaluates the correlation between edges in the ground reference and the denoised images; 
edges are detected by either the Laplacian or the Canny filter. 
This parameter ranges between $0$ and $1$, and the bigger it is, the better the filter is; ideal edge preservation yields $\beta=1$. 
PSNR is a global measure of quality, as it measures the ratio of the maximum value and the square root of the total error.
High PSNR indicates a well-filtered image.

The mean, variance and ENL are also computed within the four squares and the background.  
Good despeckling must preserve the mean value while significantly reducing the variance in these textureless areas increasing, thus, ENL.

Table~\ref{table:results_phantom_squares} presents the measures of quality as estimated in the simulated image ROIs (four squares and background), and also in the complete image.
From this table, SRAD, E-Lee and PPB performances are comparable and quite acceptable. 
However, FANS obtains most of the best scores (mainly for variance reduction and ENL) while preserving reasonably well mean values. 
MSSIM and $\beta$ are also better (for instance, $\beta=0.40$ for FANS and $\beta=0.22$ for the E-Lee filter). 
The zoom in Fig.~\ref{fig:Zoom_Phantom_Corners_All} corroborates this numerical assessment.

Table~\ref{table:results_phantom_squares} also shows the values for ${\text{ENL}}$ and the estimated $\mu$ within the background of the ratio image. 
All are close to the ideal (${\text{ENL}} \approx 1$, $\mu \approx 1$), although the best results are for FANS (${\text{ENL}} = 1.0028$ and for E-Lee ($\mu = 1.0019$).

\begin{table}[hbt]
\caption{Quantitative evaluation of filters on the simulated SAR image (best values in boldface).}
\label{table:results_phantom_squares}
\centering
\begin{tabular}{llrrrrrr}\toprule
\multicolumn{2}{l} {\textbf{Simulated SAR data}}	&\textbf{True}	&\textbf{Simulated}	&\textbf{SRAD} &\textbf{E-Lee} &\textbf{PPB} &\textbf{FANS}\\
\midrule
\multirow{3}{*}{Background}	&$\mu$	&10	&9.93	&\textbf{9.94} &9.91 &10.12	&10.13\\
& $s$&10	&9.99	&0.96	&0.92	&1.02  &\textbf{0.45}\\
& ENL	&1	&0.98	&105.86 &115.13	&90.87	&\textbf{489.38}\\
\midrule

\multirow{3}{*}{Top left square}	&$\mu$ &2	&1.96	&1.97 & 1.96 &\textbf{1.99}	&2.01\\
& $s$&2	&1.93	&0.19 & 0.19	&0.20 &\textbf{0.08}\\
& ENL	&1	&1.03	&101.80 & 106.17	& 94.64	&\textbf{640.55}\\
\midrule
\multirow{3}{*}{Top right square}	&$\mu$	&40	&40.07	&\textbf{39.98} &39.85	&40.69	&40.59\\
& $s$&40	&39.83	&4.41 & 4.24	&3.89	&\textbf{2.04}\\
& ENL&1	&1.01	&82.09 & 88.10	& 109.20	&\textbf{394.84}\\
\midrule
\multirow{3}{*}{Bottom left square}&$\mu$	&60	&59.92	&60.12 &\textbf{59.93}	&60.17	&61.54\\
& $s$&60	&60.00	&6.76	& 5.78 & 5.67&\textbf{2.88}\\
& ENL&1	&0.99	&78.88 &107.49	&112.50	&\textbf{455.83}\\
\midrule
\multirow{3}{*}{Bottom right square}&$\mu$	&80	&\textbf{79.32}	&\textbf{79.35}	&78.99 & 81.53&81.61\\
& $s$	&80	&78.89	&8.76 & 7.63	& 8.20	&\textbf{3.68}\\
& ENL	&1	& 1.01	& 81.88 & 106.99	& 96.68	&\textbf{490.45}\\
\midrule
\multirow{3}{*}{Whole image}	&{PSNR}	&---	&73.87	& \textbf{80.30} & 78.72	& 79.07	& 77.85\\
&{MSSIM}	&---	&0.38	&0.95 & 0.95	&0.95	&\textbf{0.98}\\
&{$\beta$}	&---	&0.14	&0.22 & 0.27 &0.30	&\textbf{0.40}\\
\midrule

\multirow{2}{*}{Ratio image}	& ${\widehat{{\text{ENL}}}_{\text{ratio}}}$ &1 &--- &1.0744 &1.0346 &1.0858 &\textbf{1.0028}\\
&$\widehat{\mu}_{\text{ratio}}$  &1	&--- &0.9914 &\textbf{1.0019}&0.9775 &0.9974\\
\bottomrule
\end{tabular}
\end{table}

Table~\ref{table:proposed_estimator_Phantom} shows that the proposed measure provides significantly different values for each filter.
According to $\cal M$, FANS is the best filter, followed by SRAD, E-Lee and PPB. 
The results are consistent with both the quantitative and qualitative visual assessment of the filtered images and their ratio. 
Note that FANS is the one with least geometric content within the ratio image ($\delta h = 6.26)$, and also with lowest $r_{\widehat{{\text{ENL}}}, \widehat{\mu}}$ residual.
The opposite behavior is observed in PPB, although less residual content is visible in the ratio image (compared to SRAD and E-Lee filters) it obtains the highest (worst) $\cal M$ score ($7.0371$). 
Note that this result agrees with the commonly accepted criteria of evaluation of a despeckling filter: mean and ENL must be preserved. 
Due to that high score in the $r_{\widehat{{\text{ENL}}}, \widehat{\mu}}$ residual, PPB is strongly penalized. 

\begin{table}[hbt]
\caption{Quantitative evaluation of ratio images for the simulated data (best value in boldface), computed on $n=83$ automatically detected homogeneous areas.}
\label{table:proposed_estimator_Phantom}
\centering
\begin{tabular}{l*5{r}}
  \toprule
\textbf{Filter} & $h_{\text{o}}$ &  $\overline{h_{g}}$ & $\delta h$ &$r_{\widehat{{\text{ENL}}}, \widehat{\mu}}$  & ${\cal M}$ \\ 
    \midrule
SRAD	&0.3026&0.3023&9.41&4.6634&7.0371\\
E-Lee 	&0.3465&0.3460&14.30&2.8781&8.5910\\
PPB  	&0.5551&0.5543&14.56&5.7751&10.1704\\
FANS 	&0.3827&0.3829&\textbf{6.26}&\textbf{2.0944}&\textbf{4.1816}\\
   \bottomrule
\end{tabular}
\end{table}

\subsection{Results for Actual SAR Images}

\begin{figure}[hbt]
	\centering
\subfigure{\includegraphics[width=.35\linewidth]{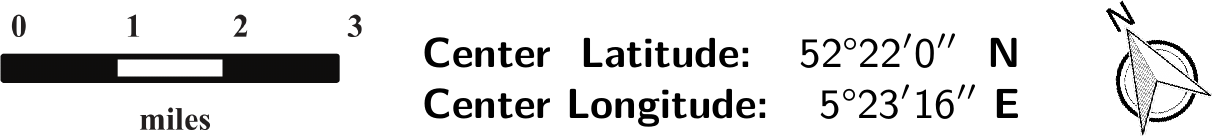}} \ \ \  	
\subfigure{\includegraphics[width=.35\linewidth]{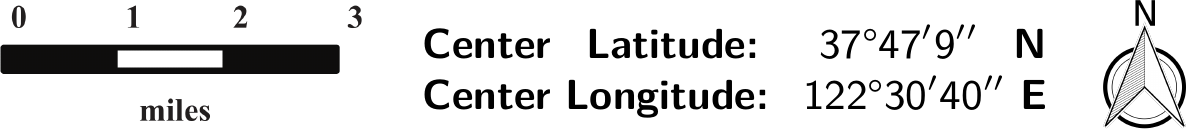}} \\	
	\subfigure[Flevoland\label{fig:AIRSARFlevoland}]{\includegraphics[width=.35\linewidth]{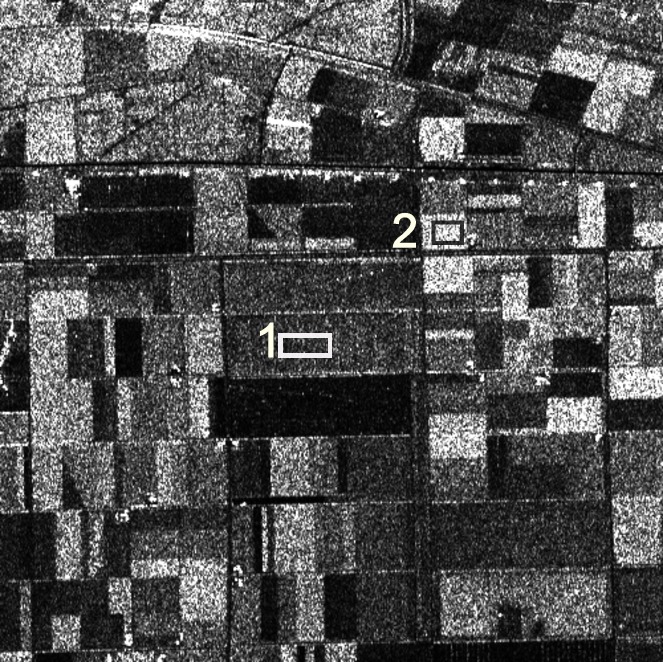}}	\ \ \ 
\subfigure[San Francisco bay\label{fig:AIRSARSanFra}]{\includegraphics[width=.35\linewidth]{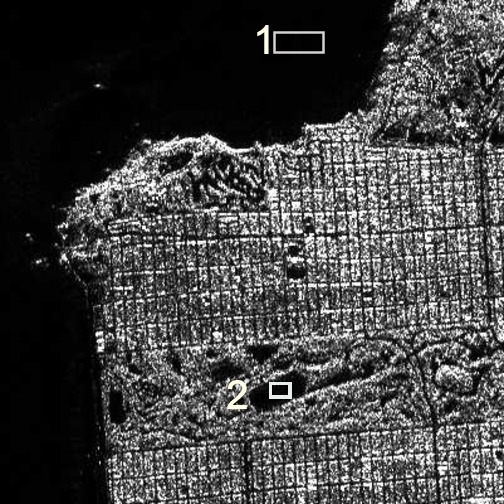}}
		\caption{Intensity AIRSAR images, HH polarization, three looks}
	\label{fig:Flevoland_SanFrancisco}
\end{figure}

We show the benefits of our proposal on two SAR images obtained by the AIRSAR sensor in HH polarization, three looks in intensity format; cf.\ Fig.~\ref{fig:Flevoland_SanFrancisco}.

\begin{figure}[hbt]
	\centering
	\includegraphics[width=.35\linewidth]{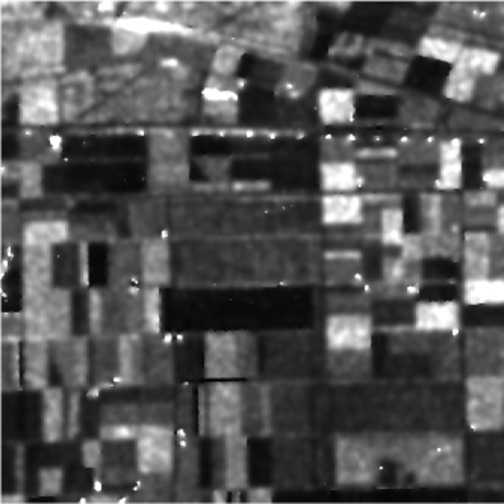}
  \includegraphics[width=.35\linewidth]{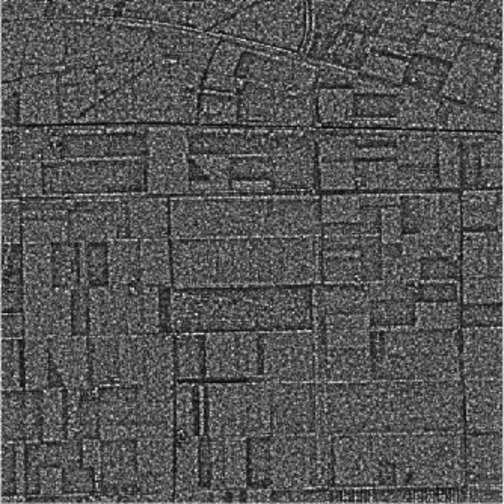}

	\vskip.5ex
	\includegraphics[width=.35\linewidth]{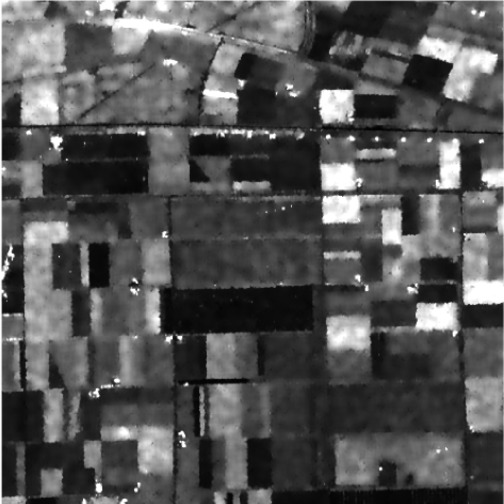}
	\includegraphics[width=.35\linewidth]{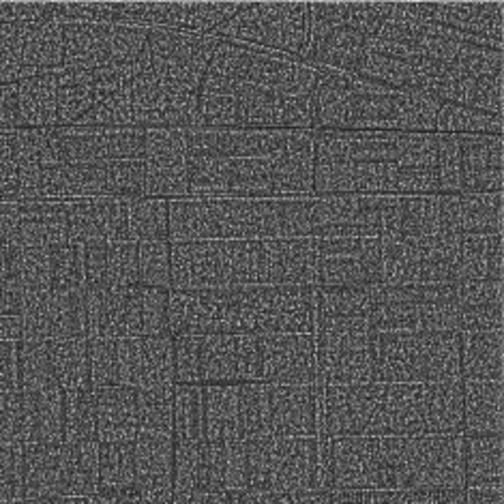}

	\vskip.5ex
	\includegraphics[width=.35\linewidth]{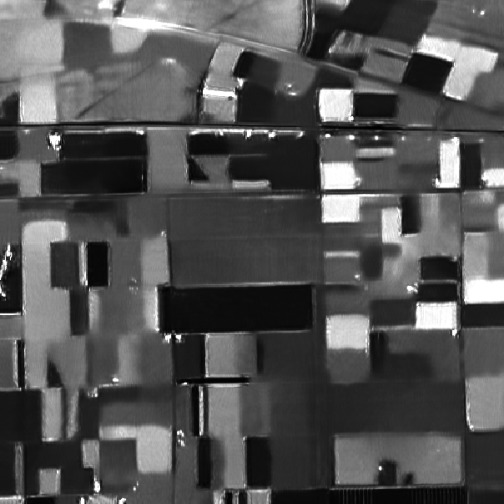}
  \includegraphics[width=.35\linewidth]{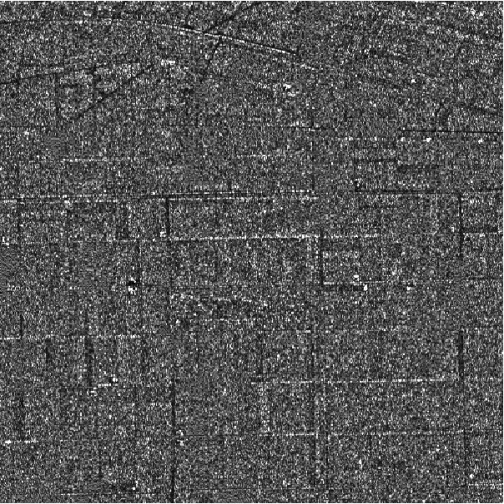}

	\vskip.5ex
	\includegraphics[width=.35\linewidth]{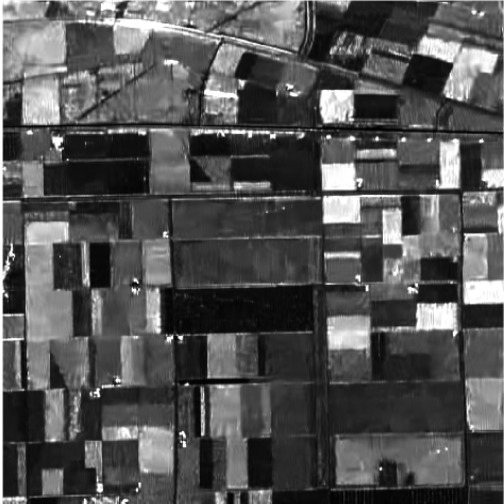}
  \includegraphics[width=.35\linewidth]{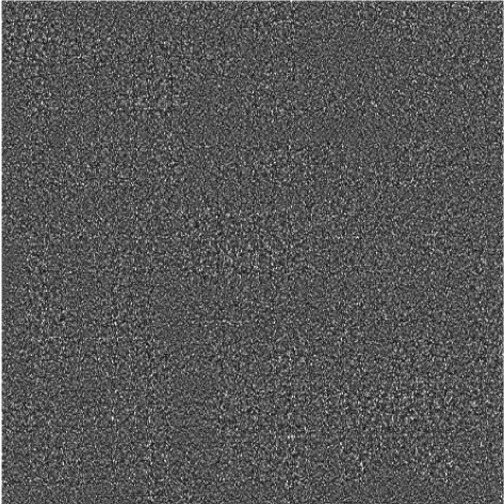}

	\caption{Results for the Flevoland image. Top to bottom, (left) results of applying SRAD, E-Lee, PPB and FANS filters. Top to bottom (right), their ratio images.}
\label{fig:Flevoland_all}
\end{figure}

Fig.~\ref{fig:AIRSARFlevoland} shows a subregion of $500\times500$ pixels from the image of Flevoland, Netherlands. 
It corresponds to a flat area made up of reclaimed land used for agriculture and forestry. 
The image contains numerous crop types grown in large rectangular fields which are very appropriate to evaluate mean and variance values. 
There are also bright scatterers which allow evaluating the filters ability at preserving them. 
Fig.~\ref{fig:Flevoland_all} shows the filtered images in the first column, and their ratio images in the second. 

\begin{figure}[hbt]
	\centering
	{\includegraphics[width=0.60\linewidth]{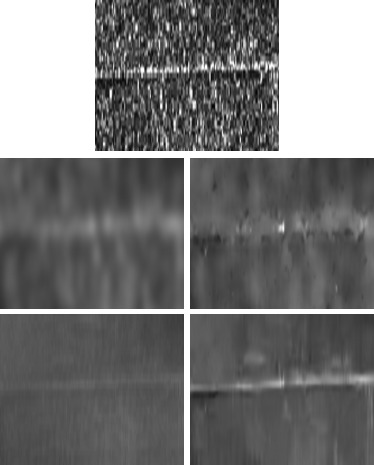}}
\caption{Zoom of the results for Flevoland image: (top) Noisy image, (first row, left) SRAD filter, (first row, right) E-Lee filter, (second row, left) PPB filter and, (second row, right) FANS filter.}
\label{fig:Flevoland_all_filters_zoom}
\end{figure}

As expected, the filters perform well in terms of variance reduction and edge and bright scatterers preservation.  
FANS (bottom) provides the best visual result, outperforming the other filters: homogeneous areas are notably more homogeneous. 
SRAD blurs a little the image. PPB gets a fine visual result but it seems also overfiltered although patch homogeneity outperforms to the other filters.
Edge preservation is better for FANS too as it can be appreciated in the images shown in Fig.~\ref{fig:Flevoland_all_filters_zoom}. 

FANS is also the best with respect to structural content in the ratio image, and SRAD is the one leaving most structure within it. 
However, as for the simulated image, minute geometrical content still remains after applying FANS. 

Table~\ref{table:estimators_Flevoland} presents the mean, standard deviation and ENL values estimated in the boxed regions identified in Fig.~\ref{fig:Flevoland_SanFrancisco} (left).
FANS is the best with respect to the mean preservation in both regions, although all filters obtain competitive values.
The best variance reduction and ENL values are obtained with PPB, notably in ROI-2. 

\begin{table}
\centering
\caption{Quantitative assessment of Flevoland filtered data in selected ROIs (best values in boldface).} 
\begin{tabular}{lrrrrrr} 
\toprule
\multirow{2}{*}{\textbf{Filter}}&\multicolumn{3}{c}{\textbf{\mbox{ROI-1}}}&\multicolumn{3}{c}{\textbf{ROI-2}}\\
\cmidrule(r){2-4}
\cmidrule(r){5-7}
&\multicolumn{1}{c}{$\widehat{\mu}$}
&\multicolumn{1}{c}{$s$}
&\multicolumn{1}{c}{ENL}
&\multicolumn{1}{c}{$\widehat{\mu}$}
&\multicolumn{1}{c}{$s$} 
&\multicolumn{1}{c}{ENL}\\ 
\midrule
Original &0.0047&0.0030&2.5000&0.0208&0.0110&3.5441\\
SRAD     &\textbf{0.0047}&7.3561e-004&41.2367& 0.0204&0.0012&283.7539\\
E-Lee    &\textbf{0.0047}&7.4516e-004&39.1870&0.0206&0.0012&276.1669\\
PPB      &0.0048&\textbf{3.3690e-004}&\textbf{200.6540}&0.0212&\textbf{5.8933e-004}&\textbf{1.2918e+003}\\
FANS    &\textbf{0.0047}&5.0309e-004&86.1449&\textbf{0.0209}&6.2295e-004&1.1290e+003\\
\bottomrule 
\end{tabular}
\label{table:estimators_Flevoland} 
\end{table}%

The analysis of the ratio images (see Table.~\ref{table:estimators_Flevoland_Ratio}) is not conclusive: no filter gets the best values for all estimators. 
PPB produced a poor ENL result in both ROI-1 and ROI-2 ($2.8048$ and $3.8159$, resp., instead of $3$). 
However, all results are acceptable with small differences and, based on the solely analysis of these estimations within the ratio images one can hardly decide if a filter performs better than the others.

\begin{table}[hbt]
\centering
\caption{Quantitative assessment of ratio images for Flevoland filtered data in selected ROIs (best values in boldface).} 
\begin{tabular}{lrrrr} 
\toprule
\multirow{2}{*}{\textbf{Filter}}&\multicolumn{2}{c}{\textbf{\mbox{ROI-1}}}&\multicolumn{2}{c}{\textbf{ROI-2}}\\
\cmidrule(r){2-3}
\cmidrule(r){4-5}
\textbf{}
&\multicolumn{1}{c}{$\widehat{\mu}$}
&ENL
&\multicolumn{1}{c}{$\widehat{\mu}$}
&ENL\\ 
\midrule
SRAD &0.9862&\textbf{2.9836}&1.0152&3.6287\\
E-Lee &\textbf{0.9981}&2.9824&\textbf{1.0097}&\textbf{3.5755}\\
PPB &0.9720&2.8048&0.9729&3.8159\\
FANS &0.9942&2.8822&0.9874&3.7082\\
\bottomrule 
\end{tabular}
\label{table:estimators_Flevoland_Ratio} 
\end{table}

In agreement with the visual inspection, FANS has the best $\mathcal M$ score (see Table~\ref{table:proposed_estimator_Flevoland}).
For these data, E-Lee obtains the worst score ($86.5863$) showing also a high $r_{\widehat{{\text{ENL}}}, \widehat{\mu}}$ residual ($11.2636$). 
It is interesting to point out that, although the best preservation of $r_{\widehat{{\text{ENL}}}, \widehat{\mu}}$ within the ratio image is provided by SRAD ($r_{\widehat{{\text{ENL}}}, \widehat{\mu}} = 8.2782$), its final $\mathcal M$ score is heavily penalized by $\delta h = 66.81$ which accounts for the remaining structural content, as expected. 
Notice that $\delta h = 1.09$ for FANS.

\begin{table}[hbt]
\caption{Quantitative evaluation of ratio images for Flevoland data (best value in boldface), computed on $n=8$ automatically detected homogeneous areas.}
\centering
\begin{tabular}{l*5{r}}
  \toprule
\textbf{Filter} 
& $h_{\text{o}}$ 
&  $\overline{h_{g}}$  
& $\delta h$
& $r_{\widehat{{\text{ENL}}}, \widehat{\mu}}$  & ${\cal M}$ \\ 
    \midrule
SRAD 	&0.2043&0.2029&66.81&\textbf{8.2782}&37.5450\\
E-Lee 	&0.2247&0.2212&161.30&11.2636&86.2863\\
PPB  	&0.6210&0.6140&114.30&10.2211&5.6174\\
FANS 	&0.8944&0.8943&\textbf{1.09}&8.8547&\textbf{4.9771}\\ 
   \bottomrule
\end{tabular}
\label{table:proposed_estimator_Flevoland}
\end{table}

In the following, we present the results for the other AIRSAR image.

Fig.~\ref{fig:AIRSARSanFra} shows a subregion of  $500\times500$ pixels from the three-look intensity AIRSAR, HH polarization, over the San Francisco Bay. 
This image contains mostly urban areas and sea, parks and hills covered by vegetation. 
There are few textureless areas except for the ocean.

\begin{figure}[hbt]
	\centering
	\includegraphics[width=.35\linewidth]{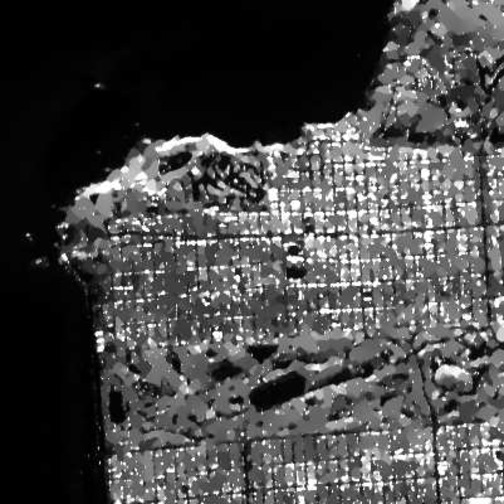}
	\includegraphics[width=.35\linewidth]{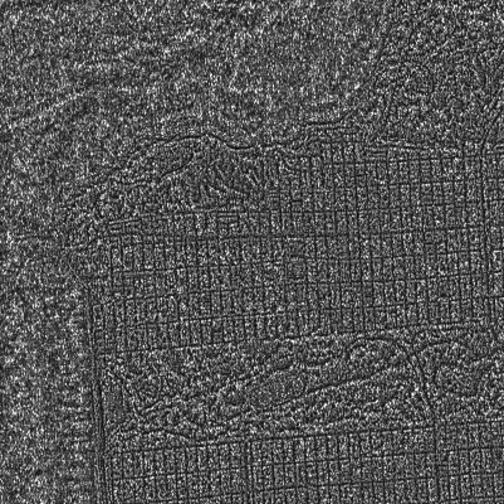}
   \vskip.5ex
   
	\includegraphics[width=.35\linewidth]{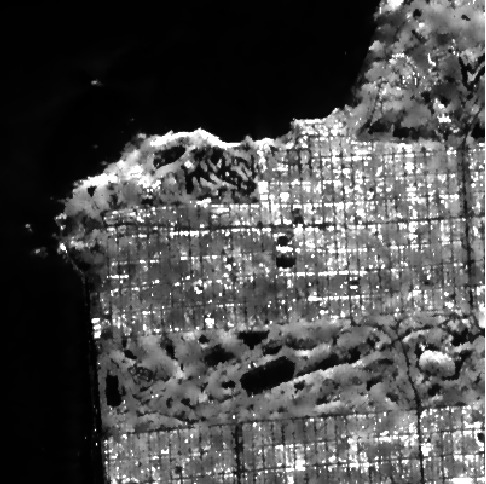}
	\includegraphics[width=.35\linewidth]{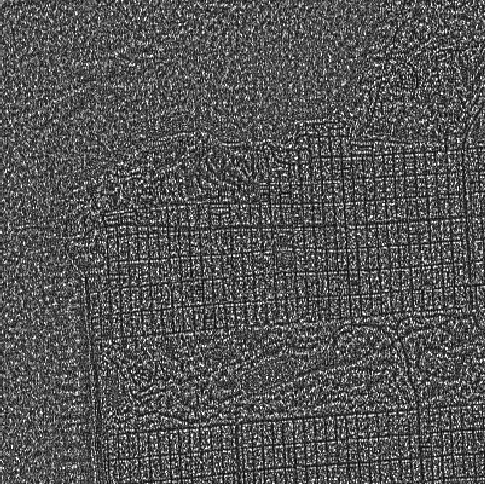}
	\vskip.5ex
	
	\includegraphics[width=.35\linewidth]{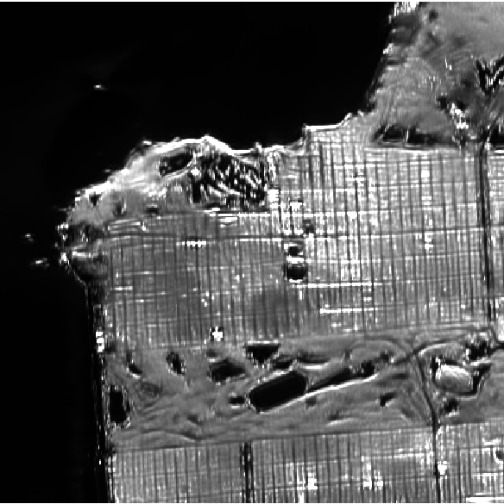}
	\includegraphics[width=.35\linewidth]{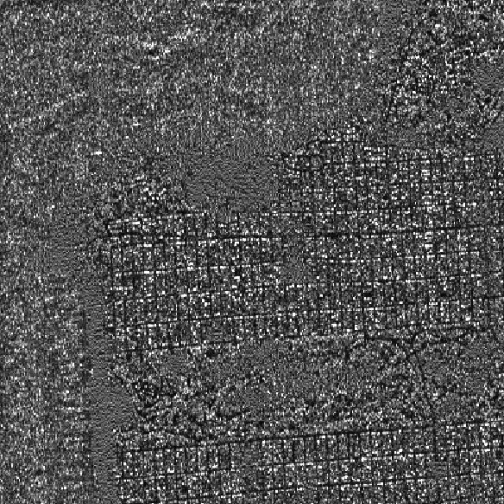}
	\vskip.5ex
	
	\includegraphics[width=.35\linewidth]{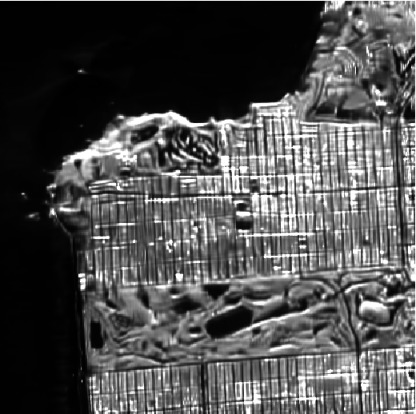}
	\includegraphics[width=.35\linewidth]{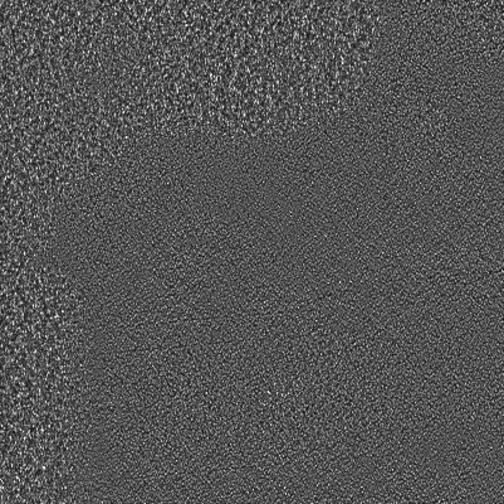}
	\caption{Result for the San Francisco bay image. Top to bottom, (left) results of applying SRAD, E-Lee, PPB and FANS. Top to bottom (right), their ratio images.}
\label{fig:SanFrancisco_all}
\end{figure}

Fig.~\ref{fig:SanFrancisco_all} presents the results obtained with SRAD, E-Lee, PPB and FANS (top to bottom, left).  
The corresponding ratio images are also shown (second column). 

SRAD clearly overfiltered and, consequently much structure is found within its ratio image. 
Notice that we have applied the recommended filter parameters~\cite{art:Yu_2002} that provided acceptable results for the simulated case and for the previous actual case (Flevoland) but, as showed, another more suitable set is required for this image. 
E-Lee preserves well the bright scatterers but parts of the image seem also overfiltered (the forest and some building blocks); as a result, much geometric content is visible in its ratio image. 
The PPB  and FANS results are visually comparable, although some bright scatterers due to buildings are lost by PPB.
FANS is also better at edge preservation.
Once again, FANS ratio image resembles pure speckle, as seen in the bottom right image, while the structural contents in the PPB ratio image are noticeable.

\begin{figure}[hbt]
	\centering
	\includegraphics[width=0.60\linewidth]{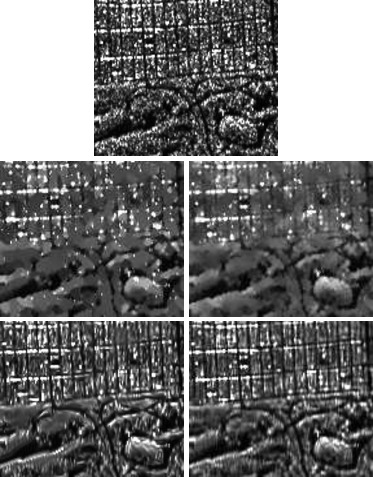}
\caption{Zoom of the results for San Francisco image: (top) Noisy image, (first row, left) SRAD filter, (first row, right) E-Lee filter, (second row, left) PPB filter and, (second row, right) FANS filter.}
\label{fig:SanFrancisco_all_filters_zoom}
\end{figure}

Fig.~\ref{fig:SanFrancisco_all_filters_zoom} shows a detail of those results. 
Notice that PPB and FANS results are visually acceptable and quite similar.

Table~\ref{table:estimators_San_Francisco_Filtered} presents the mean, standard deviation and ENL estimated in the boxed regions identified in Fig.~\ref{fig:AIRSARSanFra}. 
As with the Flevoland data, no conclusive results stem from those values. 
However, FANS is consistently the best over ROI-2. 
A similar conclusion is reached for the estimators measured on the ratio images shown in Table~\ref{table:estimators_San_Francisco_Ratio}.

\begin{table}[hbt]
\centering
\caption{Quantitative assessment of San Francisco bay filtered data in selected ROIs (best values in boldface).} 
\begin{tabular}{lrrrrrr} 
\toprule
\multirow{2}{*}{\textbf{Filter}}&\multicolumn{3}{c}{\textbf{\mbox{ROI-1}}}&\multicolumn{3}{c}{\textbf{ROI-2}}\\
\cmidrule(r){2-4}
\cmidrule(r){5-7}
&\multicolumn{1}{c}{$\widehat{\mu}$}
&\multicolumn{1}{c}{$s$}
&\multicolumn{1}{c}{ENL}
&\multicolumn{1}{c}{$\widehat{\mu}$}
&\multicolumn{1}{c}{$s$} 
&\multicolumn{1}{c}{ENL}\\ 
\midrule
Original &6.8327e-004&3.8422e-004&3.1625&0.0018&8.9834e-004&4.1959\\
SRAD &7.0597e-004&3.6942e-005&365.2071&0.0021&8.1671e-005&674.1768\\
E-Lee &\textbf{6.8252e-004}&8.9163e-005&58.5959&\textbf{0.0020}&1.7975e-004&129.6569\\
PPB &6.9884e-004&\textbf{3.2459e-005}&\textbf{463.5443}&\textbf{0.0020}&1.5831e-004&163.9516\\
FANS &6.9156e-004&4.7095e-005&215.6278&\textbf{0.0020}&\textbf{3.5513e-005}&\textbf{3.2947e+003}\\
\bottomrule 
\end{tabular}
\label{table:estimators_San_Francisco_Filtered} 
\end{table}

\begin{table}[hbt]
\centering
\caption{Quantitative assessment of ratio images for San Francisco bay filtered data in selected ROIs (best values in boldface).} 
\begin{tabular}{lrrrr}
\toprule
\multirow{2}{*}{\textbf{Filter}}&\multicolumn{2}{c}{\textbf{\mbox{ROI-1}}}&\multicolumn{2}{c}{\textbf{ROI-2}}\\
\cmidrule(r){2-3}
\cmidrule(r){4-5}
\textbf{}
&\multicolumn{1}{c}{$\widehat{\mu}$}
&\multicolumn{1}{c}{ENL}
&\multicolumn{1}{c}{$\widehat{\mu}$}
&\multicolumn{1}{c}{ENL}\\
\midrule
SRAD &0.9651&\textbf{3.3477}&0.8634&4.7106\\
\hline
E-Lee &\textbf{0.9955}&3.5834&0.8948&4.9673\\
\hline
PPB &0.9692&3.4437&0.9004&4.7289\\
\hline
FANS &0.9829&3.3819&\textbf{0.9023}&\textbf{4.2443}\\
\bottomrule 
\end{tabular}
\label{table:estimators_San_Francisco_Ratio} 
\end{table}

Table~\ref{table:proposed_estimator_San-Francisco} presents the proposed $\cal M$ metric. 
Again, FANS obtains the best score as expected from the visual inspection of the ratio images. 
Although the best result for ${\text{ENL}}$ value and $\mu$ preservation is for the SRAD filter, due to the large amount of residual structure within its related ratio image, $\delta h$ is large enough to rank it to the last position among all despeckling filters discussed in this work.

\begin{table}[hbt]
\caption{Quantitative evaluation of ratio images for San Francisco bay data (best value in boldface), computed on $n=10$ automatically detected homogeneous areas.}
\label{table:proposed_estimator_San-Francisco}
\centering
\begin{tabular}{l*5{r}}
  \toprule
\textbf{Filter} & $h_{\text{o}}$ &  $\overline{h_{g}}$  & $\delta h$ & $r_{\widehat{{\text{ENL}}}, \widehat{\mu}}$  & ${\cal M}$ \\ 
    \midrule
SRAD	&0.5643	&0.5368	&487.26	&\textbf{0.2216}	&5.0942\\
E-Lee 	&0.5813	&0.5586	&390.35	&0.3262	& 4.2297\\
PPB  	&0.7449	&0.7419	&40.65	&0.5395	& 0.9460\\
FANS 	&0.7138	&0.7141	&\textbf{5.10}	&0.4231	&\textbf{0.4741}\\  
      \bottomrule
\end{tabular}
\end{table}

The above results for actual SAR data support the use of our proposed $\cal M$ metric.

\subsection{Using $\cal{M}$ for filter design}
\label{Sec:Design}

Next we show the use of $\cal{M}$ in fine-tuning the parameters of a despeckling filter on actual data. 
We use FANS due to its already attested performance, and the Niigata Pi-SAR data as the image to be despeckled.

\begin{figure}[hbt]
	\centering
	\subfigure{\includegraphics[width=.3\linewidth]{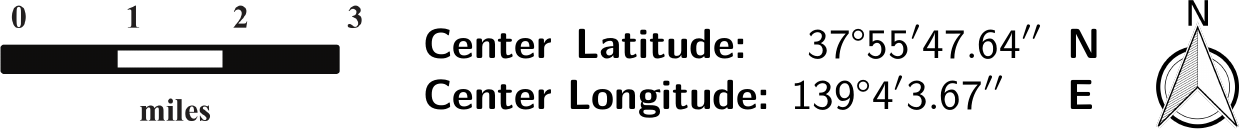}} \vspace{-0.25cm}
	 \\	
	\subfigure{\includegraphics[width=.3\linewidth]{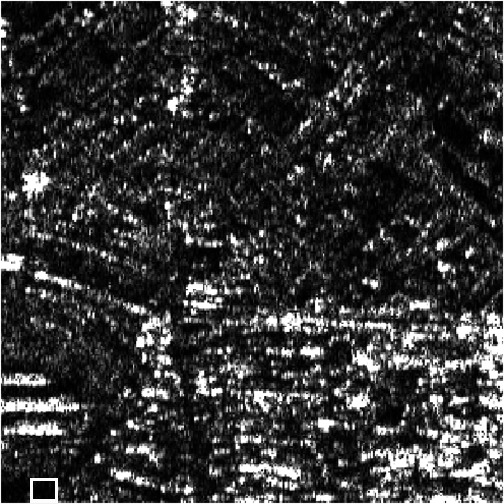}}	 
		\subfigure{\includegraphics[width=.3\linewidth]{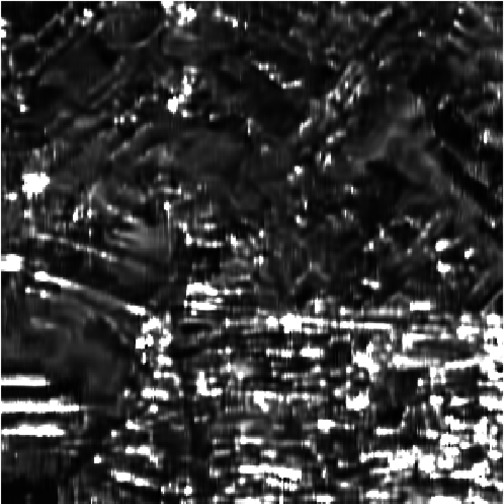}}	
	\subfigure{\includegraphics[width=.3\linewidth]{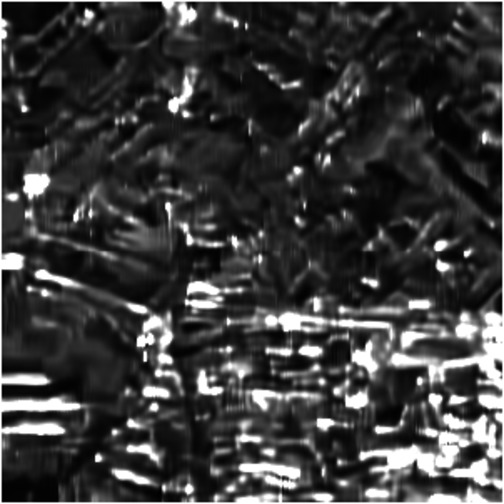}}
	\caption{Intensity Pi-SAR, HH one look Niigata image (left). Results of applying FANS filters with default parameters (middle) and with optimized parameters (right).}
\label{fig:Niigata_Fans}
\end{figure}

Fig.~\ref{fig:Niigata_Fans} (left) shows a subimage ($300 \times 300$ pixels), in intensity format, one look and HH polarization. 
The resolution of this image is \SI{3x3}{\squared\meter}. 
The selected area includes urban and forest patches.

As indicated in~\cite{art:Cozzolino_2014} FANS requires more than ten control parameters, although the authors also mentioned that ``\textit{All parameters have been set once and for all, obtaining always satisfactory results in the experiments, so the user can forget about them and keep the default values}". 

However, we show that some improvement can be achieved by a basic optimization strategy. 
We selected three control parameters: 
$S$ (size of rows and columns of neighborhood blocks), 
$P_{FA}$ (false alarm probability related to wavelet thresholding for the classification process), and
 $W$ (wavelet transform used in the 2D spatial domain). 
 The default values for these parameters are $S = 16$, $P_{FA} = 10^{-3}$ and, the Daubechies-4 wavelet for the choice of $W$. 
 These control parameters are extensively discussed in~\cite{art:Cozzolino_2014}, and they seem to have a strong impact on the filter performance.

The filter was optimized by exhaustive search: 
$S\in[4,20]$ with steps $h_{S} = 1$,
$P_{FA}\in[0.001, 0.01]$ with steps $h_P = 0.001$, and 
wavelet transforms from the ones suggested in the author's Matlab implementation: Meyer, DCT (discrete cosine transform), Haar,  Daubechies-2, Daubechies-3, Daubechies-4, biorthogonal-1.3, and biorthogonal-1.5. 

The optimal values found were $S = 4$, $P_{FA} = 0.0041$ and the Haar wavelet transform.
With these, eighteen $15 \times 15$ homogeneous areas were detected.

\begin{figure}[hbt]
	\centering
   \includegraphics[width=.3\linewidth]{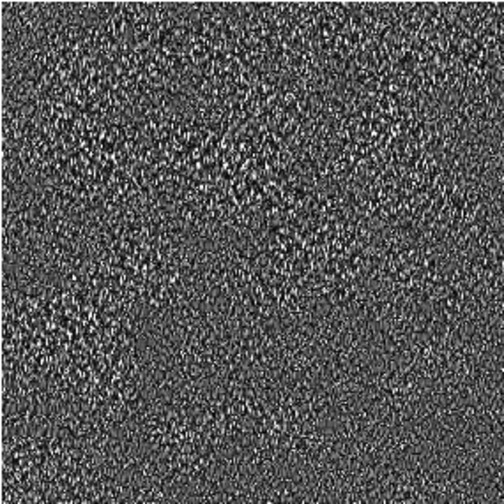}
   \includegraphics[width=.3\linewidth]{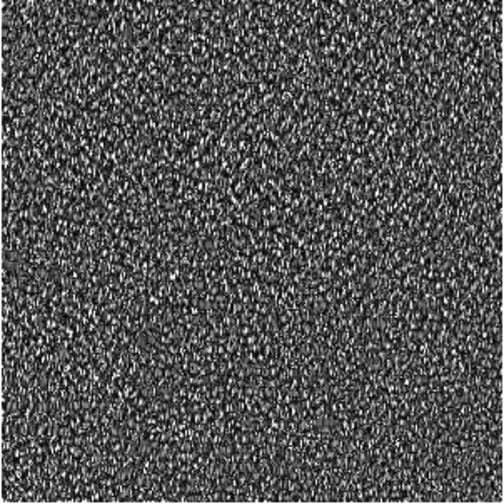}

	\caption{Ratio images for Niigata data (left: FANS with default parameters and with optimized parameters (right).}
\label{fig:Niigata_Fans_Ratio}
\end{figure}

The despeckled result by FANS with default parameters is shown in Fig.~\ref{fig:Niigata_Fans} (middle) and the result by using the optimized parameters is shown in the same figure (right). 
A seen, some artifacts have been notably reduced and homogeneous areas, which seem more uniform with the optimized filter. 
The ratio images are depicted in Fig.~\ref{fig:Niigata_Fans_Ratio}. 
A visual inspection suggests that there remains less geometrical structure within the ratio image by filtering with the optimized parameters.  

%
%

Table~\ref{table:estimators_Niigata} presents the mean, standard deviation and ENL estimated in the boxed region identified in Fig.~\ref{fig:Niigata_Fans} (left). Best results for the three estimators are for the optimized FANS filter.

 \begin{table}[hbt]
\centering
\caption{Quantitative assessment of San Francisco bay filtered data in selected ROIs (best values in boldface).} 
\begin{tabular}{lrrr}
\toprule
\textbf{Filter} & 
\multicolumn{1}{c}{$\widehat\mu$} & 
\multicolumn{1}{c}{$s$} & 
\multicolumn{1}{c}{ENL} \\ 
    \midrule
Original &0.0283&0.0261&1.1757\\
FANS (default parameters) &0.0302&0.0106&8.1171\\  
FANS (optimized parameters) &\textbf{0.0295}&\textbf{0.0083}&\textbf{12.6325}\\  
      \bottomrule
\end{tabular}

\label{table:estimators_Niigata} 
\end{table}
 
Similar conclusion is reached for the estimators measured on the ratio images shown in Table~\ref{table:Measures_Niigata_Ratio}.

\begin{table}[hbt]
\caption{Quantitative evaluation of ratio images for Niigata data (best values in boldface), computed on $n=18$ automatically detected homogeneous areas.}
\centering
\begin{tabular}{lcc}
  \toprule
\textbf{Filter} & $\widehat\mu$ & ENL \\ 
    \midrule
FANS (default parameters) &0.8627&2.2270\\  
FANS (optimized parameters) &\textbf{0.9006}&\textbf{1.8745}\\  
      \bottomrule
\end{tabular}
\label{table:Measures_Niigata_Ratio}
\end{table}

The proposed $\cal M$ metric is presented in Table~\ref{table:proposed_estimator_Niigata}. 

\begin{table}[hbt]
\caption{Quantitative evaluation of ratio images for Niigata data (best value in boldface), computed on $n=18$ automatically detected homogeneous areas.}
\label{table:proposed_estimator_Niigata}
\centering
\begin{tabular}{l*3{r}}
  \toprule
\textbf{Filter}  & $r_{\widehat{{\text{ENL}}}, \widehat{\mu}}$ & $\delta h$  & ${\cal M}$ \\ 
    \midrule
FANS (default parameters)  &0.4833 &20.89 &10.6867\\ 
FANS (optimized parameters)  &\textbf{0.3794} &\textbf{6.50} &\textbf{3.4397}\\ 
      \bottomrule
\end{tabular}
\end{table}

From these results, it is clear that $\cal M$ can be applied to design a despeckling filter working on actual data without the need of ground references.

\section{Conclusions}
\label{Sec:Conclusion}

We proposed a new image-quality index, $\cal M$, to objectively evaluate despeckling filters. 
The proposal operates only in the ratio image and requires no reference.
The evaluation relies on measuring deviations from the ideal statistical properties of the ratio image and their residual structural contents.
The last component is computed by comparing a textural measure in the ratio image with random permutations of the data.

We have shown the expressiveness and adequacy of $\cal M$ using both simulated data and SAR images, and we verified that it is consistent with widely used image-quality indices as well as with the visual inspection of both filtered and ratio images.
It has been also shown that the proposed unassisted image quality index can also be embedded into the design of despeckling filters.
Additionally, the computational cost related to the proposed estimator is comparable to state-of-the art indexes.

The proposal is valid as long as the multiplicative model holds and provided that at least one (even small) region can be detected as textureless.
The user is required to input an estimate of the number of looks.
The index employs a random component, but it is reproducible once fixed the platform, the random number generator, and the seed.

\vspace{6pt} 

\supplementary{The code and data for reproducing the results here reported are available here \url{http://www.de.ufpe.br/ ~raydonal/ReproducibleResearch/UNASSISTED/UNASSISTED-QUANTITATIVE.html}.
Computational platform and running times are informed in the Appendix.}

\acknowledgments{The third author (ACF) wishes to express his gratitude to Prof.\ Xin Li (Chinese Academy of Sciences, Lanzhou) for the two-month stay during which this work was concluded.}

\authorcontributions{All authors contributed equally to this manuscript.}

\conflictsofinterest{The authors declare no conflict of interest. 
The founding sponsors had no role in the design of the study; in the collection, analyses, or interpretation of data; in the writing of the manuscript, and in the decision to publish the results.} 

\appendixtitles{yes} 
\appendixsections{one} 
\appendix
\section{Computational platform}
The Matlab~\cite{Matlab_2014} language was used to simulate and analyze the data. Haralick's textural features were also computed by the available libraries in Matlab.
The computational cost for the $500 \times500$ synthetic data shown in this work (see Fig.~\ref{fig:Phantom_corners}) with the parameters setting as mentioned in Section~\ref{Sec:Experimental Setup and Results}, is around \SI{20}{\second} in an Intel Core i7 Q740 \SI{1.73}{\giga\hertz} machine.

\section{Extension to the Gaussian additive noise model}
The idea of using the residual image as a proxy for filter quality can be also used for the Gaussian additive model.
If the observed image is $Z=X+Y$, with $X$ and $Y$ independent fields, and $Y$ a collection of iid zero-mean Gaussian random variables, then the ideal filter will produce $\widehat X=X$, and the residual image $I=Z-\widehat X=Y$ should bear no structure and be formed by Gaussian deviates with zero mean and the same variance.
This idea was used by Hale~\cite{StructureOrientedBilateralFiltering} to attest to the superiority of a new filter for seismic images.
The analysis is visual, so there is room for research using, for instance, the Anderson-Darling test for normality.
Peng et al.~\cite{UnifiedFrameworkRepresentationBasedSubspaceClustering} also analyze residuals as a measure of quality of subspace clustering.


\bibliographystyle{mdpi}


\end{document}